\documentclass[11pt,a4paper]{article}
\usepackage{times,latexsym}
\usepackage{url}
\usepackage[T1]{fontenc}

\usepackage{authblk} 

\usepackage[acceptedWithA]{tacl2018v2}

\usepackage{xspace,mfirstuc,tabulary}

\newif\iftaclinstructions
\taclinstructionsfalse
\iftaclinstructions
\renewcommand{\confidential}{}
\renewcommand{\anonsubtext}{(No author info supplied here, for consistency with
TACL-submission anonymization requirements)}
\newcommand{\instr}
\fi

\iftaclpubformat 

\else

\fi

\usepackage{graphicx}
\usepackage{tabularx}
\usepackage{algorithm}
\usepackage{algorithmic}

\title{Measuring prominence of scientific work in online news as a proxy for impact}

\author[1,3,4]{James Ravenscroft}
\author[2]{Amanda Clare}
\author[1,3]{Maria Liakata}

\affil[1]{Centre for Scientific Computing, University of Warwick, CV4 7AL, United Kingdom}
\affil[2]{Department of Computer Science, Aberystwyth University, SY23 3DB, United Kingdom}
\affil[3]{Alan Turing Institute, 96 Euston Rd, London, NW1 2DB, United Kingdom}
\affil[4]{Filament AI, 22 Tudor St, London,  EC4Y 0AY, United Kingdom}

\date{}
\begin{document}
\maketitle
\begin{abstract}
  The impact made by a scientific paper on the work of other academics has many established metrics, including metrics based on citation counts and social media commenting. However, determination of the impact of a scientific paper on the wider society is less well established. For example, is it important for scientific work to be newsworthy? Here we present a new corpus of newspaper articles linked to the scientific papers that they describe.  
  We find that Impact Case studies submitted to the UK Research Excellence Framework (REF) 2014 that refer to scientific papers mentioned in newspaper articles were awarded a higher score in the REF assessment. The papers associated with these case studies also feature prominently in the newspaper articles.
  We hypothesise that such prominence can be a useful proxy for societal impact. We therefore provide a novel baseline approach for measuring the prominence of scientific papers mentioned within news articles. Our measurement of prominence is based on semantic similarity through a graph-based ranking algorithm. We find that scientific papers with an associated REF case study are more likely to have a stronger prominence score. This supports our hypothesis that linguistic prominence in news can be used to suggest the wider non-academic impact of scientific work.
\end{abstract}

\section{Introduction}
Understanding the comprehensive impact of scientific work is motivating for academics and helpful for demonstrating value to research funding councils. For example, the UK government conducts an evaluation of academic research known as REF (Research Excellence Framework), which asks universities to submit impact case studies evidencing the \textit{comprehensive} (non-academic) impact of a selection of their research \cite{Ref20142011}. This impact may be societal, environmental, health, policy-related or financial. REF impact case studies are awarded a score, that, together with other measures, is used to decide how to distribute core funding. As academics, we would like to improve our impact and to understand where and how that can arise.

Beyond centrally coordinated schemes such as REF, measuring comprehensive impact remains difficult due to the large number of ways work could have an impact (e.g. new medical procedures, patents, spin off businesses) and a lack of available data. Although not all impactful work is reported in the news, traditional news outlets provide a key source of information for members of the general public wishing to learn about ongoing scientific research \cite{MacLaughlin2018}. A newspaper article may describe a new finding (such as a gene linked to a disease), the rise of a popular research area (such as self-driving cars), or a business venture based on a scientific discovery. We are interested to discover whether news coverage of scientific work can be used as a proxy for the comprehensive impact of that work and whether the prominence of the scientific work in news articles plays a role.

The text of the REF impact case studies from the most recent exercise (2014) is available online and can be analysed automatically using natural language processing. The case studies have been awarded a score on a scale from unclassified, denoting that a scientific work has had little reach or impact, to 4*, which denotes outstanding reach and significance of impact \cite{Ref20142011}.In this work we build a new corpus of linked news articles, scientific papers, and REF impact case studies, and use it to test the hypothesis that case studies with scientific papers that are mentioned in the news obtain a higher REF impact score.

As a next step we consider the way in which news articles mention the scientific work. For example, is it just in passing, is it part of a wider article about the state of a field, or is the scientific work in question the main component of the article? To achieve this we define a measure of  \textit{prominence} in news articles, and we develop a method to determine whether a scientific paper is prominently featured in the news article. 
Furthermore, we apply a publicly available\footnote{\url{http://www.sapientaproject.com/}} system for scientific discourse segmentation~\cite{liakata2012sapienta} to help us characterise the scientific contribution of each sentence. This allows us to break down a scientific paper into its main component parts (background, goals, methods, outcomes etc.) and inspect those that contribute most to the news article. This paves the way for future work into automated analysis of the knowledge transfer from scientific work to newspaper articles. We expect this will allow scientists to understand how they might better engage the media to have more impact. 


Specifically the contributions of this paper are as follows:
\begin{enumerate}
    \item We provide a new corpus of newspaper articles, scientific papers and REF 2014 impact case studies and the relation between them.
    \item We show that the REF impact case studies that have one or more linked news articles are likely to have been awarded a higher impact score than case studies that are not linked to news articles.
    \item 
    We describe and evaluate a novel baseline approach for measuring the prominence of a scientific work within a newspaper article. 
    \item We find that scientific papers used in REF case studies have a greater prominence in their linked newspaper articles than the papers that were not used in REF case studies. We therefore conclude that prominence of a scientific paper in a news article is a useful proxy measure for impact that goes wider than academia. 
    \item We finally explore which parts of a scientific paper (Goals, Background, Methods, Outcomes) feature prominently in a news article.

\end{enumerate}



\section{Corpus of linked news articles and scientific papers}
\label{sec:corpus}
We gathered news content from a range of 1.3 million broadsheet and tabloid digital newspaper articles from UK outlets: BBC News, The Guardian, The Telegraph, The Daily Express, The Independent and The Daily Mail. The majority of these articles were taken from the JISC web archive which covers articles between 1996 and 2013\cite{JISC2013}. The Guardian content spans 1921-2018. We searched for news articles that mentioned scientific articles and we assembled a corpus of 5903 digital news articles and 9891 linked scientific papers. Out of the latter we were able to find the full text for 1097 of them with only meta-data available for the rest (e.g. DOI, title and names of authors). While a lot of work on analysing the scientific literature is performed with abstracts, we require the full scientific article as many findings of interest to the news are not reported in the abstracts.

\begin{figure}
    \centering
    \includegraphics[width=0.9\columnwidth]{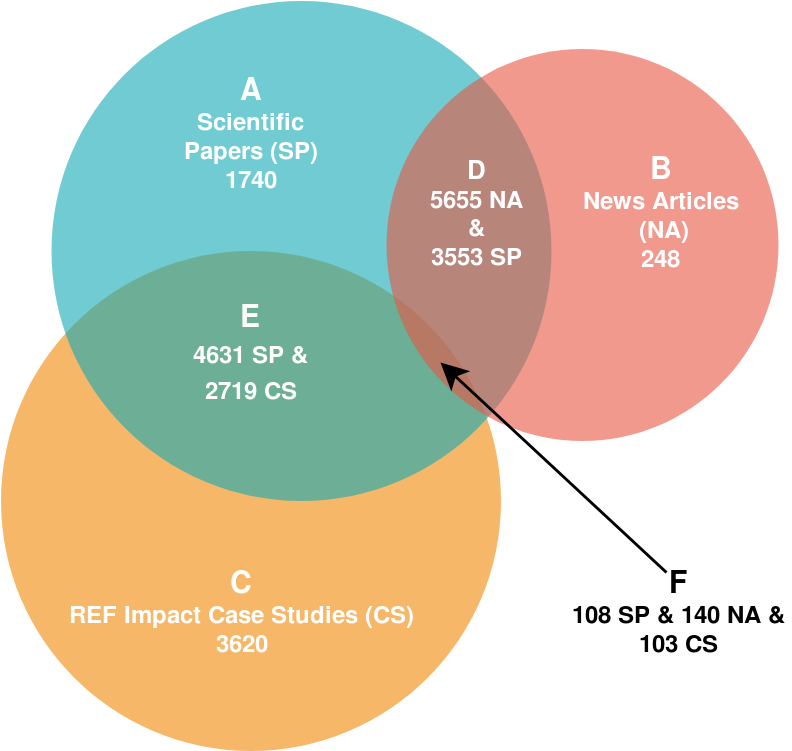}
    \caption{Documents in the corpus according to size and relation.}
    \label{fig:corpus_venn}
    \vspace{-0.5cm}
\end{figure}

A newspaper article is considered to be linked to a scientific work if it explicitly mentions or references said work within its content. Broadsheet format papers such as The Guardian and BBC News often directly cite scientific work either as a traditional citation or by providing a hyperlink or DOI leading to the digital location of the work. In these cases, we were able to use an automated approach to record links. Regular Expressions were used to match DOI strings in newspaper text. For extracting hyperlink references, we developed a web scraping script to follow the links and extract DOIs from the HTML metadata tags embedded in the target web pages. These metadata tags are  mostly standardised across journal publishers. We used gathered metadata to query the Unpaywall web service \cite{Piwowar2018} in order to find full text content for 1097 of the scientific papers (see Figure \ref{fig:corpus_venn}). We attribute the larger number of linked scientific papers compared to the news articles mentioning them to the fact that news articles have a tendency to reference more than one scientific work.

Tabloid format papers such as The Daily Express and The Daily Mail tend to make passing or implicit references to scientific work, normally omitting DOIs and URLS and usually in the format ``Researchers at $<Institution>$ have published a study in $<journal>$". In these cases, we made use of HarriGT, an open source tool that is able to suggest and infer links using named entities, keywords and dates found in the article text and their similarity to scientific paper metadata \cite{Ravenscroft2018}. 

It was also possible to link a subset of the corpus to REF case studies that mention either an associated news article or scientific paper. Historical REF case studies are freely available to browse and read via a web portal\footnote{\url{https://impact.ref.ac.uk/CaseStudies/CaseStudy.aspx?Id=5794}}. We downloaded all 6640 case studies for the 2014 REF assessment and then used a script to identify outbound DOIs and hyperlinks in the case studies. We then cross referenced these outbound relationships with our existing corpus of scientific papers. We also considered outbound hyperlinks from REF case studies to news articles in our corpus. In total, we were able to identify 103 case studies with links to 140 news articles and 108 scientific papers (see (F) in Figure \ref{fig:corpus_venn}). Our linked corpus is included as a supplementary material with this submission.

With the subset of our linked corpus that has REF impact case study information, we were able to explore relationships between REF impact and news coverage.

\section{Methodology}

\subsection{From Newspaper Coverage to Comprehensive Impact}
\label{sec:newspaper_coverage_impact}

\begin{figure}
    \centering
    \includegraphics[width=0.9\columnwidth]{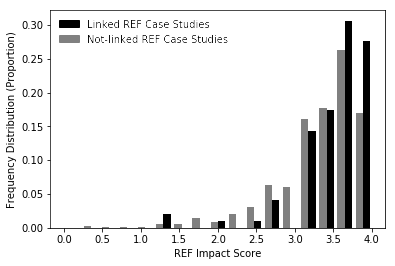}
    \caption{Frequency Distribution of REF Impact Scores for REF case studies that are linked ($F \cup G$, blue) or not linked ($C \cup D$,red) to one or more news article.}
    \label{fig:popularity_distribution}
    \vspace{-0.5cm}
\end{figure}

REF impact scores for individual case studies are not published in order to preserve anonymity of academics. Instead, the number of 4*, 3*, 2*, 1* and unclassified case studies are reported per unit of assessment (UoA, approximately: faculty or department) at each partaking institution. 

REF requires a minimum of two impact case studies to be submitted per UoA with an additional case study per 10 Full Time Employees (FTE) \cite{Ref20142011}. We found that the mean number of FTE per participating scientific UoA was 27.3 and thus the mean number of case studies submitted per UoA was 3-4.

Since a more granular impact score cannot be obtained, we follow the approach of Ravenscroft et al. \shortcite{Ravenscroft2017}, assigning the mean impact score of the case studies from the associated UoA and institution as the score for individual case studies. Although some information loss is inevitable, in 96\% of submissions the mean calculation involved 10 or fewer data points. 

REF impact scores are very important to all UK universities and very few impact case studies may be submitted per UOA. There is also a 4 page limit on all impact case studies. Therefore, UK universities are heavily incentivised to select their best scientific work for inclusion into case study submissions at UoA level. We presume that the inclusion of a scientific paper in a REF case study is an indicator of high comprehensive impact. The vast majority of scientific papers are not included in REF case studies either due to the small number of impact case studies submitted per institution UoA, the timing of the papers in relation to REF or the authors being external to the UK. 

We split all REF impact case study results into two sets depending on whether they had news articles linked to them. We refer to the subset of REF case studies that have a link to one or more newspaper articles and optionally also one or more scientific papers in our corpus as ``linked" (F in Figure \ref{fig:corpus_venn}) and the remaining REF case studies that do not have any known relationship to news articles or scientific papers as ``not linked" (C in Figure \ref{fig:corpus_venn}). 
Figure \ref{fig:popularity_distribution} shows a plot of the frequency distribution for these these two sets against average REF impact score as per the calculation above.

Our hypothesis is that scientific papers mentioned in newspaper articles ($D \cup F$) have a higher comprehensive impact than those that are not. We first used D'Agostino and Pearson's normality test \cite{Dagostino1973} and found that neither set of scores has a normal distribution ($p=8.66\times10^{-17}$ and $p=1.01\times10^{-220}$ for papers with and without mentions in the news respectively). We therefore opted to use the non-parametric Kolmogorov-Smirnov 2-sample test \cite{Massey1951} (KS-2 Test) to test the significance of the difference between the two distributions.  
The KS-2 test shows that the two samples are most likely drawn from separate populations ($p=0.007$). A two-sample bootstrap test of mean difference \cite{Hesterberg2015} gives error bounds of [0.07,0.27] suggesting that REF impact case studies with one or more associated news article have a higher impact score than those that are not linked to news articles. Having established that a relationship between news mentions and REF impact score exists, we seek to understand the extent to which the context of the mention of a scientific article within a news article affects REF impact.

\subsection{Prominence in News Articles}
\label{sec:prominence}


We hypothesise that scientific papers which are discussed prominently in a news article are likely to generate more comprehensive impact than scientific works that are mentioned in passing. For example a news article discussing the growth of AI in the finance industry may refer to multiple scientific papers introducing technology, but the core message of the article may be about human jobs rather than the science.

Prominence refers to the importance assigned to text by its author. It is a somewhat under investigated area of computational linguistics and even traditional linguistics and discourse \cite{BECKER201828}. Prominence has recently received some attention in the domain of argumentation mining with Wachsmuth et al. \shortcite{Wachsmuth2017a} taking the position that prominence may be considered ``a product of popularity" rather than a measure of intrinsic quality or importance. 

In the same spirit Boltuzic and \v{S}najder \shortcite{Boltuzic2015} focus on repetition as a key indicator of prominence in order to automatically identify common arguments in online debates. They use clustering to group semantically similar arguments together before manually analysing and labelling clusters. This approach gives some insight into prominence of arguments found in online debates but does not offer any narrative on the intrinsic importance of the arguments presented.

In contrast to prominence,  salience corresponds to the intrinsic importance of a unit of text within a document regardless of its presentation. Boguraev \shortcite{Boguraev1997} uses salience as a way to measure the ``aboutness" of a document. 

LexRank \cite{Erkan2004} and TextRank \cite{Mihalcea2004} are two popular and related methods for extractive summarisation that use graph centrality as a way to understand the salience of a text unit such as a sentence within a document. In TextRank and the continuous variant of LexRank, documents are represented as a fully connected graph where each sentence is a vertex and edges represent the semantic similarity between two sentences. Both approaches use a ranking approach based on PageRank \cite{Page1999} to identify those sentences which are most central to the document graph and prioritise them for inclusion in the summary.

Although prominence and salience can be defined and used in their own right, there is often overlap between the prominence and salience of a unit of text since the most important details in a document are often those that the author is trying to communicate most clearly. However there are some cases where salience and prominence are not aligned. In an advertisement for a credit card, the small-print about interest rates and terms of repayment provides salient information about the product but is not normally presented prominently. Conversely, ``clickbait" headlines are often prominent in order to capture the attention of the reader but often misrepresent the article they are attached to, thus having low salience.

Examining prominence and salience together can help us understand how important and informational the units of text are within a document and how this compares to the way that information is presented by an author. 

We posit that beyond salience, TextRank and LexRank both take into account and prioritise prominent information too. LexRank and TextRank scoring is based on the similarity of each sentence with respect to all others, where similarity is defined respectively in each paper. It therefore follows that information repeated in multiple sentences (and thus prominent in the document) will boost the relative similarity of these sentences and thus the score of all sentences that discuss the repeated information.

We use a modified version of continuous LexRank \cite{Erkan2004} to create a method for measuring and ranking all sentences in news articles from the Linked Corpus in terms of the prominence of the information they contain. We call this method ``SemSimRank''. 

\begin{figure}[h]
\centering
\includegraphics[width=3cm]{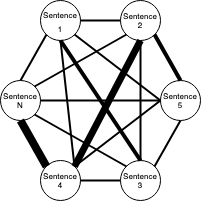}
\caption{A fully connected, undirected, weighted graph representing the pairwise semantic similarity of sentences in a document. Edge weight is equal to the semantic similarity between vertices.}
\label{fig:weighted_sentences}
\vspace{-0.5cm}
\end{figure}

For a given news article document $D$ containing $S$ sentences, pairwise semantic similarity $\theta(s_i, s_j)$ between all sentences $s_i; i\in\{0.. S\}$ and $s_j; j \in \{0..S; j \neq i \}$ is stored in adjacency matrix $E$ and used to create a fully connected weighted graph $G(D,E)$ as per Figure \ref{fig:weighted_sentences}. Edges are normalised row-wise to help with ranking convergence. 

We then use PageRank \cite{Page1999} with damping factor $\alpha$, max iterations $N$ and convergence threshold $\delta$ to produce a set of rankings $P$ for all sentences $s_i \in D$\cite{Mihalcea2004, Erkan2004}. This process is described in Algorithm \ref{algo:SemSimRank}.

The key difference between SemSimRank and TextRank and LexRank approaches is how pairwise semantic similarity $\theta(s_i, s_j)$ is calculated. LexRank uses the cosine similarity of TF-IDF vectors\cite{Erkan2004} whilst TextRank simply uses the word overlap between the two sentences \cite{Mihalcea2004}. We evaluate a number of semantic similarity functions $\theta(s_i, s_j)$ based on text representations explored in section \ref{sec:semantic_similarity} in order to understand the information transfer between the most prominent sentences in the news articles and excerpts of the linked scientific papers. For each article we select $N$ top sentences for comparison with the scientific content.

\begin{algorithm}
\caption{SemSimRank Sentence Selection Algorithm}
\label{algo:SemSimRank}
\begin{algorithmic}
\FOR{$i$ in [0..$S$]}
\FOR{$j$ in [0..$S$]}
 \STATE{$E_i^j = \theta(s_i, s_j)$}
\ENDFOR

\STATE{$E_i =  \frac{E_i}{\sum\limits_{j=0}^{S}E_i^j}$}
\ENDFOR

initialise $P_i=1/S$ for $i$ in $S$ 

\WHILE{num iterations $< N$}
  \FOR{$i$ in [0..$S$]}
  \STATE{$P^{new}_i = \frac{1-d}{S} + d\sum\limits^S_{j=0}{\frac{P_j}{\sum\limits_{z=0}^{S}E_j^z}}   $}
  \IF{$\sum^{S}_{i}(P^{new}_i-P_i) < \delta$}
  \STATE {return $P^{new}$}
  \ENDIF
  \STATE{$P = P^{new}$}
  \ENDFOR

\ENDWHILE
\end{algorithmic}
\end{algorithm}

\subsection{Prominence in Scientific Papers}


The structure of scientific papers is very different from that of a newspaper article. Scientific papers are typically much longer and much more complex than news articles and are written for a specialised audience rather than a general lay audience. It is therefore not appropriate to apply methods, such as SemSimRank used above, that work well for newspaper articles directly to scientific papers without some adaptation\cite{Teufel2002}. To our knowledge there is no previous work that investigates linguistic prominence within scientific papers. However, the domain of scientific discourse which focuses on identifying the functional role of units of text within scientific articles is well explored and provides a useful starting point for understanding prominence within scientific work. 

Argumentative Zoning \cite{Teufel1999,Teufel2010} (AZ) is a seminal piece of work in this area which defines an annotation scheme for rhetorical structures within scientific papers. AZ aims to ``...[capture] the attribution of intellectual ownership in scientific articles, expressions of authors' stance towards other work, and typical statements about problem-solving processes.'' (\textit{ibid.})
Variants of AZ have been applied to both scientific articles and article abstracts. Subsequently, Liakata et al. \shortcite{liakata2010} introduced an annotation scheme complementary to AZ, the Core Scientific Concept (CoreSC) annotation scheme. CoreSC aims to capture the content and structure of a scientific investigation rather than its rhetorical narrative and related arguments. 

CoreSC and AZ provide complementary metadata that is useful for a wide range of tasks including summarisation of scientific papers \cite{Teufel2002,Liakata2013}, information retrieval \cite{Teufel2006, Duma2016} and prediction of a publication's communication style \cite{Ravenscroft2013}.  As per our definition in Section \ref{sec:prominence} we consider prominence to be intrinsically tied to the importance assigned to a text by its author. We hypothesise that the functional information provided by scientific discourse annotations is also useful for determining the prominence of sentences therein and that certain discourse categories are likely to be more prominent than others. For example, authors are likely to emphasise their novel contributions and findings by featuring them clearly in their conclusion.

The authors of AZ and CoreSC advocate for their combined use to leverage their individual strengths \cite{liakata2010}. In our work, we utilise CoreSC annotations owing to the presumed relation between CoreSC content-based categories and our definition of prominence and to its larger training corpus and publicly available automated SAPIENTA classifier \cite{liakata2012sapienta}. The latter can be used via a web service\footnote{\url{http://sapienta.papro.org.uk/}}. It is also advantageous that SAPIENTA is trained primarily on biomedical papers which make up the  majority of our linked scientific content due to journalists favouring these kinds of papers \cite{MacLaughlin2018}. We first use SAPIENTA to assign CoreSC scientific discourse labels (e.g. all Results, Hypotheses, etc.) to each sentence in the scientific article corpus linked from the newspaper articles. 

Some types of CoreSC scientific discourse categories are quite rare (e.g. Hypotheses, Objectives) and therefore we aggregate the 11 discourse categories together into 4 CoreSC Groups: Background (containing Background and Motivation), Goals (containing Goals, Objectives and Hypotheses), Method (containing Method, Experiment and Model) and Outcomes (containing Observations, Results and Conclusions).

As a next step we measure the semantic similarity between the extracts from the news articles that are considered prominent by the SemSimRank algorithm and the linked scientific articles, considering a pairwise similarity with each sentence extracted from the scientific paper. 

To measure semantic similarity we use a number of metrics discussed below in Section \ref{sec:semantic_similarity}.  We then take the mean semantic similarity for sentences assigned to each CoreSC category. This allows us to find the extent to which the content apportioned to each CoreSC category is encapsulated by the news excerpt.
\vspace{-0.3cm}

\subsection{Semantic Similarity}
\label{sec:semantic_similarity}

We aim to identify a sensible baseline approach for determining the semantic similarity between the news excerpts that are prominent and CoreSC passages extracted from linked scientific works such that it is possible to measure knowledge transfer between the documents. We compare a number of common feature representations and similarity metrics in order to find the most suitable for this task. This is especially important since most semantic similarity tasks such as STS \cite{agirre2012} and SICK-R \cite{marelli2014} involve short documents (typically sentences) rather than comparison between long documents.

Initially we use bag-of-words (BoW) count vectors to represent both scientific papers and newspaper articles. We combine a static English stopwords list, a lowercase filter and a minimum word length check of 3 characters or more in order to identify sets of relevant unigram features. The combined corpus vocabulary is very sparse. However, since we only consider pairwise semantic similarity between linked documents we need not attempt to build a global feature model for the corpus, avoiding problems associated with a high dimensional feature space. Instead we construct a local feature space on-the-fly for each document pair allowing us to avoid feature pruning and retain all valid unigram features within specific document pairs. Using count vectors rather than simpler binary one-hot encoding allows us to account for word repetition, which may be indicative of prominence, within our similarity calculations.

BoW feature representations are unable to account for semantic relationships between distinct words. Therefore approaches like the one described above may fail to successfully encapsulate relationships between two related documents that use different vocabularies with minimal word overlap. Even when discussing the same subject matter, scientific papers and newspaper articles are typically written in different grammatical styles and vocabularies for scientists and laypeople respectively. This motivates us to consider alternative sentence representations that are more sensitive to semantic relationships between texts.

Semantic vector space representations such as GloVe \cite{Pennington2014} and word2vec \cite{Mikolov2013} aim to capture semantic similarity between distinct words by taking into account the context in which they are used via word co-occurrence matrices and neighbouring words within a sentence respectively. Words are assigned a real-valued vector within a multi-dimensional semantic space and semantically similar words are assigned vectors with a strong cosine similarity.  


In our task we employ pre-trained GloVe\footnote{\url{https://nlp.stanford.edu/projects/glove/}} 
feature embeddings trained on the Common Crawl dataset\footnote{\url{http://commoncrawl.org/}}, a multi-petabyte archive of content scraped from the world wide web containing 42 billion tokens and a vocabulary 1.9 million words. 
Sentences from the document pairs are represented by taking the mean of word vectors associated with each individual token for a given excerpt, an approach recently shown to yield state-of-the-art performance for sentence similarity/matching tasks \cite{Shen2018}. 

GloVe embeddings are not sensitive to word use in context which is problematic when words have multiple meanings. Furthermore, GloVe is unable to provide good representations of rare and previously unseen words resulting in suboptimal representations of sentences containing words that were not in the training corpus. To address these problems, we additionally evaluate BERT, a recent neural model that produces context sensitive embeddings of words and sentences\cite{devlin2018bert}. BERT also attempts to address the `out-of-vocabulary' problem by capturing and encoding subword embeddings which can be summed together on the fly to generate vectors for missing words\cite{devlin2018bert}. In our experiment we make use of the freely available pre-trained BERT model\footnote{\url{https://github.com/google-research/bert}}. The relative merits of varying neural embedding approaches are further discussed in Section \ref{sec:related_work} below. We use BERT to encode documents sentence-by-sentence and compare the generated sentence vectors.

We use cosine similarity for measuring the similarity between semantic embeddings-based sentence vectors from pairs of news articles and scientific papers. For our BoW count vector representation we use Jensen-Shannon Distance (defined as square root of Jensen-Shannon divergence) instead of cosine similarity since, the former is more suited to sparsely populated non-normalised integer word count vectors.

\section{Results and Discussion}




Given that UK universities will be submitting their best work to the REF assessment process, we assume that our collection of 140 scientific works (collection F, Figure \ref{fig:corpus_venn}) associated with both a newspaper article and REF case study will have a distribution of comprehensive impact levels skewed towards higher impact than the wider collection of papers mentioned in news articles (collection D - F, Figure \ref{fig:corpus_venn}).

We evaluate pairwise semantic similarity for excerpts from all known linked pairs of newspaper articles and scientific papers using both F and D paper collections and each of the above-described feature representations and respective similarity metrics. A statistically significant uplift in semantic similarity between prominent science and news excerpts from the fully linked collection F versus the collection D would indicate that our notion of prominence is related to the comprehensive impact of scientific papers featured in the news.

\begin{figure*}[ht]
    \vspace{-0.7cm}
    \centering
    \includegraphics[width=\textwidth]{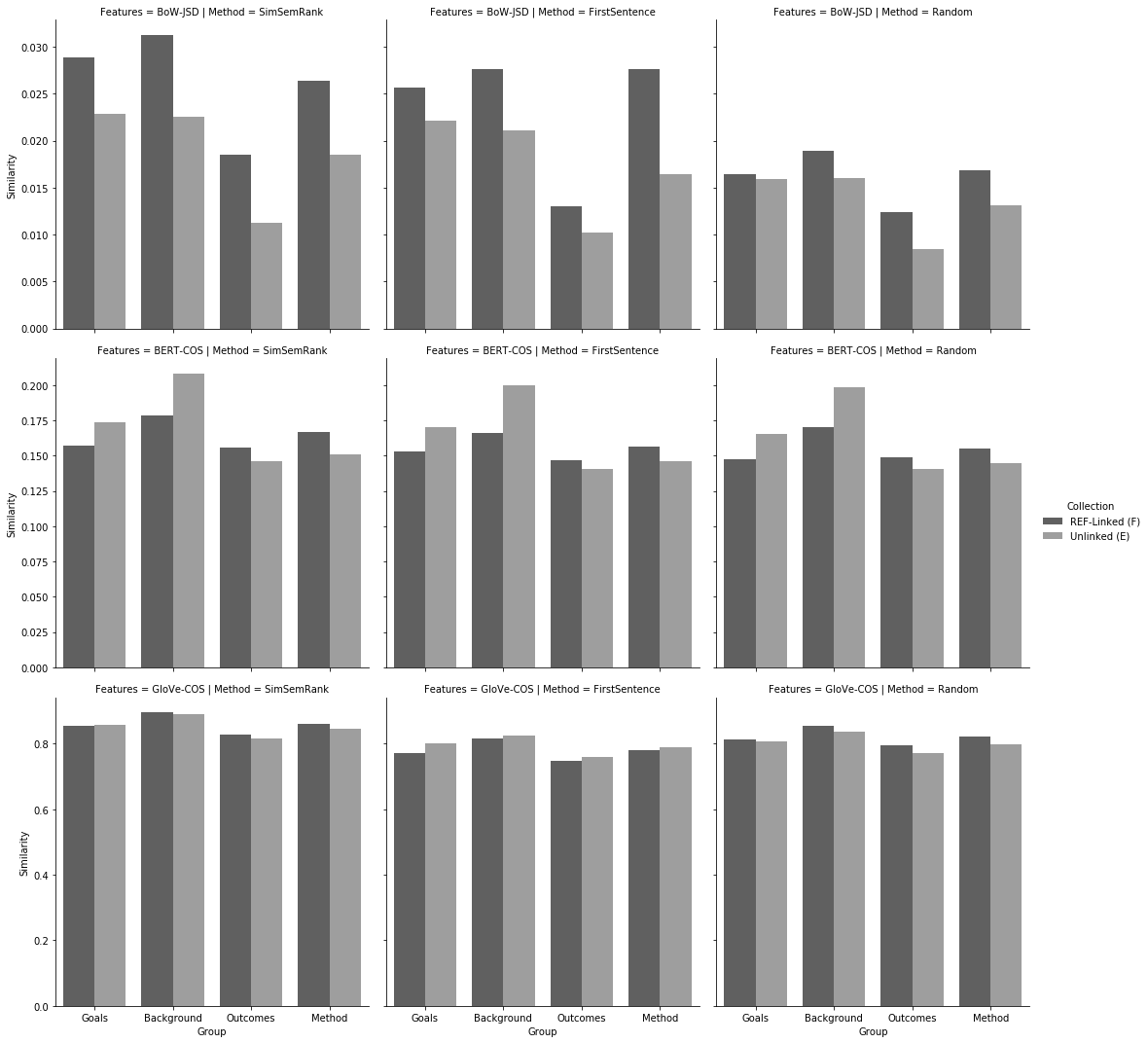}
    \vspace{-0.3cm}
    \caption{Mean Pairwise \% Difference in Semantic Similarity between newspaper articles and scientific papers that are associated with REF case studies versus those that are not (orange). Scales are not comparable across rows.}
    \label{fig:results_grouped}
\end{figure*}

We use our SemSimRank algorithm to select the most prominent sentence from each newspaper article for pairwise comparison against each sentence from linked scientific papers. We also measure our SemSimRank algorithm against two baseline approaches:

Our ``First Sentence" baseline simply takes the first full sentence from each newspaper article as the most prominent sentence. Since news paper articles often start with an overview of their content, this is a simple but often effective strategy. Our ``Random Sentence" baseline uniformly at random selects one of the sentences from the newspaper article as the most prominent sentence. We preserve the random choice across all experiments so that the results can be compared.

In most cases, the pairwise sentence comparisons yield a low number of highly similar sentence pairs and a large majority of sentences that have a low semantic similarity. Given that the median of these similarity distributions is often close to zero, we report the mean similarity in our results. We also found that the smoothing effect of averaging our results was preferable over taking the maximum value which is prone to noise from outliers.

\begin{table*} 
\hspace{-1cm}
\vspace{-3mm}
\small
\begin{tabular}{|p{8.5cm}|p{8.0cm}|}
\hline
\textbf{News Article - First Sentence}                                                         & \textbf{News Article - SemSimRank Sentence}                                                                                      \\\hline
It isn't often that science and pop culture overlap, but the two fields are in agreement when it comes to the familiar trope of the forgetful stoner. &  But with the recent changes in drug policy, the chances are that more people will be smoking cannabis than ever before, and the more potent and more popular high-THC/low-CBD marijuana that is available today will increase their risk of dependence.                                                                                                                   \\\hline
Writing in the British Medical Journal they say a 15\% cut in consumption could save 8.5 million lives around the world over the next decade.  &  The report - by researchers at the Universities of Warwick and Liverpool - says that after cutting tobacco consumption, getting people to eat less salt would be the most cost effective way to improve global health. \\ \hline
Several prehistoric creatures developed elaborate body traits in order to attract members of the opposite sex, according to new research.  & Co-author Dr Dave Martill from the University of Portsmouth said: "Pterosaurs put even more effort into attracting a mate than peacocks whose large feathers are considered the most elaborate development of sexual selection in the modern day".
\\
\hline
\end{tabular}
\caption{Example pairs of First Sentences and corresponding ``most prominent'' sentences discovered by SemSimRank from the same articles.}
\label{tab:prominent-sentences}
\end{table*}

Our results are shown in Figure \ref{fig:results_grouped}. For each CoreSC group we show the observed mean semantic similarity between REF-linked documents in Collection F and non-REF-linked in Collection D. Our findings suggest that for these collections scientific work is more prominently discussed in REF-linked newspaper articles than non-REF-linked articles. For the 'Outcomes and 'Methods' CoreSC groups in particular, almost all of the experiments in Figure \ref{fig:results_grouped} show stronger similarity for the REF-linked documents than the unlinked documents. However, both the feature set and newspaper sentence ranking approach have a strong impact on how well this relationship is captured.

All BoW-JSD approaches seem to consistently capture a significant positive difference in prominence for REF-linked documents across all 4 CoreSC discourse groups. The most effective method is BoW-JSD + SemSimRank  but BoW-JSD + FirstSentence also captures this difference particularly well (both pass KS-2 Test with $p<0.05$). The relative success of the FirstSentence approach may be down to the large number of news articles within the linked corpus that begin with a first sentence briefly summarising the key goals and outcomes from the linked scientific document.  However, there are also a significant number of articles that do not start in this way, engaging readers in a more chatty, informal style (see Table \ref{tab:helpful-firstsent}). In these instances, BoW-JSD + SemSimRank typically outperforms BoW + FirstSentence by identifying a more relevant summary sentence in the newspaper article. Table \ref{tab:prominent-sentences} shows examples of FirstSentence and SemSimRank selected sentences from the same article.

\begin{table*}[h!]
\small
\begin{tabular}{|p{7.5cm}|p{7.5cm}|}
\hline
\textbf{News Article First Sentences - High Semantic Similarity}                                                                     & \textbf{News Article First Sentences - Low Semantic Similarity}                                                     \\
\hline
One in three adults aged over 65 in England have difficulty understanding basic health-related information, suggests a study in the BMJ.         & Like many patient groups, the Alzheimers' Society isn't happy with the state of scientific research.. \\
\hline
Acne drug not found to increase suicide risk                                                                                         & It has all the makings of a pub quiz teaser: what do Barack Obama, Emma Watson, Jake Gyllenhaal and the British TV presenter Fiona Bruce have in common?                        \\
\hline
University College London researchers found a 3.6\% decline in mental reasoning in women and men aged 45-49.
&
Lately, it seems as if everyone is anti-antidepressants.\\






\hline

\end{tabular}
\caption{Example `First Sentence' extracts that are helpful (left) and not helpful (right) for prominence task }
\label{tab:helpful-firstsent}
\end{table*}

Figure \ref{fig:results_grouped} also shows that methods employing semantic embeddings do not perform as well at this task as BoW methods. All three ranking methods paired with BERT-COS features tend to show stronger semantic similarity for the Unlinked collection in `Goals' and `Background'. The ranking methods also seem to have little effect on the similarities generated by the BERT-COS comparison approach. The GloVe-COS approach is the least consistent of the methods, generating semantic similarities that are barely discernable for the two document collections.

For GloVe, we hypothesise that the lack of embeddings for uncommon and unseen words from these methods makes them less suited to this task which is very dependent on rare and specific words like names of academics, institutions, methodologies and instruments that may only occur in one pair of linked documents. BERT attempts to infer embeddings for missing tokens by combining subword information using Wu et al\shortcite{Wu2016} WordPiece model implementation \cite{devlin2018bert}. However, Wu et al\shortcite{Wu2016} suggest that subword models are not particularly helpful at representing entity names and numbers which are important features in our task.

\section{Related Work}
\label{sec:related_work}

\textbf{Scientific impact} Interest in new scientific impact metrics has grown in recent years, catalysed by government funding providers' increasing interest in scientific outputs beyond the scientific paper and subsequent citations (e.g. supporting datasets, software and subsequent social media activity \cite{Piwowar2013},  societal impact of scientific work \cite{Ref20142011, Lane2010}). Metrics aiming to quantify these alternative outputs via online activities (`altmetrics') are increasingly recognised tools \cite{Warren2017}, exhibiting specific strengths and weaknesses \cite{Ortega2018} for measuring specific outputs. Our work focuses on societal impact of science, which recent research has shown is not reflected in popular altmetrics \cite{Bornmann2019}. Maclaughlin et al. \shortcite{MacLaughlin2018} investigate linguistic features that make scientific work likely to be covered by news articles. However, our paper takes the discussion of the relationship between science and news further in order to understand what sort of news coverage is indicative of scientific impact.

\textbf{Semantic Embeddings} Neural embedding approaches popularised by word2vec \cite{Mikolov2013} and GloVe \cite{Pennington2014} provide a baseline feature set that can be used to build static representations of words. More recent systems such as ElMo \cite{Peters2018} and BERT\cite{devlin2018bert} provide promising out-of-the-box word representations sensitive to word usage and context. Likewise, BERT is also able to learn to generate vectors that encode a whole sentence, setting the current SOTA performance benchmark on a range of NLP tasks \cite{devlin2018bert}. For longer documents, such as those in our corpus, methods like BERT which focus on word and sentence level embeddings are not effective without the document segmentation and hierarchical feature representation discussed above. Works such as multi-depth attention-based hierarchical recurrent neural networks \cite{Jiang2019SemanticTM} and Rotational Unit of Memory \cite{Dangovski2019} have also shown that useful vector representations of long documents are attainable and that they can be used for NLP tasks involving long sequences. This could allow us to explore better ways to find prominent excerpts of scientific works in future.

\textbf{Semantic Similarity}
Detection of semantic similarity is a well defined task with popular annual workshops \cite{agirre2012, marelli2014}. Current state of the art models  successfully measure semantic similarity within the context of these purpose-built corpora\cite{Subramanian2018}. However, semantic similarity between newspaper articles and scientific papers is a very different task with more lexical dissimilarities between the documents being compared. Recent works that could facilitate comparison of lexically different, semantically similar documents include Kutuzov et al \shortcite{Kutuzov2018} who model shifts in word meaning over time and Conneau et al.\shortcite{Conneau} who align semantic embedding models trained on completely different languages without supervision.

\section{Conclusion}

We have provided a novel linked corpus of news articles and scientific papers, together with evidence of their impact from REF. A new measure of prominence has been defined and evaluated. We conclude that, whilst not all high impact scientific papers generate press interest, the prominence of scientific papers that are mentioned in news articles is a useful proxy measure for comprehensive impact.

We would also like to conduct further work to investigate whether  learning to align embedding models trained on news and science respectively \cite{Conneau, Kutuzov2018} could improve our performance at the prominence task.

We have made our code and datasets freely available for future research on github (URL here after anonymous review).


\bibliography{paper}
\bibliographystyle{acl_natbib}

\end{document}